\documentclass[conference]{IEEEtran}
\IEEEoverridecommandlockouts
\usepackage{cite}
\usepackage{amsmath,amssymb,amsfonts}
\usepackage{algorithmic}
\usepackage{subcaption}
\usepackage{graphicx}
\usepackage{textcomp}
\usepackage{multirow}
\usepackage{xcolor}
\usepackage{booktabs}
\def\BibTeX{{\rm B\kern-.05em{\sc i\kern-.025em b}\kern-.08em
    T\kern-.1667em\lower.7ex\hbox{E}\kern-.125emX}}
\begin{document}
\title{A Cooperation Graph Approach for \\ Multiagent Sparse Reward Reinforcement Learning \\
{\footnotesize 
}
}

\author{
\IEEEauthorblockN{Qingxu Fu\IEEEauthorrefmark{1}\IEEEauthorrefmark{2},
Tenghai Qiu\IEEEauthorrefmark{1}\IEEEauthorrefmark{2},
Zhiqiang Pu\IEEEauthorrefmark{1}\IEEEauthorrefmark{2},
Jianqiang Yi\IEEEauthorrefmark{1}\IEEEauthorrefmark{2},
Wanmai Yuan\IEEEauthorrefmark{3}
}
\IEEEauthorblockA{\IEEEauthorrefmark{1}\textit{University of Chinese Academy of Sciences}, Beijing, 100049, China
}
\IEEEauthorblockA{\IEEEauthorrefmark{2}\textit{Institute of Automation}, \textit{Chinese Academy of Sciences}, Beijing, 100190, China }
\IEEEauthorblockA{\IEEEauthorrefmark{3}\textit{Electronics Technology Group Corporation} \textit{Information Science Academy of China} Beijing, China \\
qingxu.fu@outlook.com, tenghai.qiu@ia.ac.cn, \\ zhiqiang.pu@ia.ac.cn, 
jianqiang.yi@ia.ac.cn, yuanwanmai7@163.com
}
}


\maketitle

\begin{abstract}
Multiagent reinforcement learning (MARL) can solve complex cooperative tasks. 
However, the efficiency of existing MARL methods relies heavily on well-defined reward functions.
Multiagent tasks with sparse reward feedback are especially challenging
not only because of the credit distribution problem, 
but also due to the low probability of obtaining positive reward feedback.
In this paper, 
we design a graph network called Cooperation Graph (CG).
The Cooperation Graph is the combination of two simple bipartite graphs,
namely, the Agent Clustering subgraph (ACG) and the Cluster Designating subgraph (CDG).
Next, based on this novel graph structure,
we propose a Cooperation Graph Multiagent Reinforcement Learning (CG-MARL) algorithm,
which can efficiently deal with the sparse reward problem in multiagent tasks.
In CG-MARL, 
agents are directly controlled by the Cooperation Graph.
And a policy neural network is trained to manipulate this Cooperation Graph,
guiding agents to achieve cooperation in an implicit way.
This hierarchical feature of CG-MARL provides space
for customized cluster-actions,
an extensible interface for introducing fundamental cooperation knowledge.
In experiments, 
CG-MARL shows state-of-the-art performance in sparse reward
multiagent benchmarks, including the anti-invasion interception task 
and the multi-cargo delivery task.


\end{abstract}


\section{Introduction}

The success of reinforcement learning greatly inspired research in multiagent systems.
Different from classic single-agent problem such as Go \cite{silver2016mastering} and
RTS game \cite{vinyals2017starcraft, vinyals2019grandmaster}.
multiagent reinforcement learning (MARL) focuses on creating cooperative policy for multiple agents,
e.g. Dota 2 \cite{berner2019dota}, SMAC \cite{usunier2016episodic}, 
multi-robot encirclement \cite{zhang2020multi}
and attacker-defender \cite{wu2021multi}.
However, by carefully tuning environment rewards,
all these tasks and solutions tactfully avoid the sparse reward problem, 
which leads to great difficulty not only in MARL 
but also in classic single-agent RL problems.

There are many reasons that the sparse reward problem must not be treated as a trivial issue in MARL.
Firstly,
reward-tuning is likely to introduce biases 
that mislead agents to learn unexpected cooperative behaviors.
Secondly,
the reward re-designer must be knowledgeable about the task itself
and have enough resources for the environment reverse-engineering.
Thirdly,
reward hand-crafting tricks for one problem are likely to fail in the other one
and are valuable to reward poisoning \cite{rakhsha2021reward}.

Compared with single-agent sparse reward challenges \cite{vinyals2019grandmaster},
MARL sparse reward challenges are fundamentally different in sparsity.
In single-agent problems, 
RL algorithm will eventually achieve decomposition of rewards in the dimension of time (delayed rewards \cite{watkins1989learning}),
in which way the key actions leading to success are reinforced. 
However, rewards are decomposed in the dimension of individuals (for credit assignment) in MARL
\textbf{before} they can be decomposed in the dimension of time.
To highlight this distinction between them,	
we define \textbf{first-order sparsity} (1st-sparsity) to describe the difficulty of single-agent sparse reward problems,
and \textbf{second-order sparsity} (2nd-sparsity) to describe that of MARL sparse reward problems.

Until now,
tools capable of learning sparse-reward multiagent problems are limited.
A practical solution is exploring general intrinsic rewards \cite{achiam2017surprise}, such as curiosity mechanism \cite{still2012information, groth2021curiosity}.
For some specific multiagent tasks,
it is possible to use curriculum learning \cite{gupta2017cooperative, bengio2009curriculum} to train multiagents starting from a very small scale,
then gradually expand the size of the team.
As another alternative,
imitation learning \cite{reddy2019sqil, vecerik2017leveraging, le2017coordinated} from a huge expert database is an efficient option for resourceful organizations.
But such an approach is restricted by regulations protecting data privacy \cite{pan2019you} and depends heavily on human accuracy.

An additional problem is the limited computational resource.
High-performance servers are expensive, and
agents are usually trained in a simplified environment to improve efficiency.
Some obvious cooperative behaviors are impossible to emerge in RL due to these simplifications.
For example, geese use V-shaped formation in migration to reduce energy costs. 
To learn such formation, 
we need to use energy cost as one of the reward terms and run aerodynamics calculations, 
which might take days to simulate seconds of just a single episode.
This fact reminds us that a framework that
combines emergence agent behaviors with human's prior knowledge about cooperation is absent.
Constructing such a framework can significantly 
reduce the cost of learning the most obvious pattern of cooperation 
when dealing with tasks of real-world relevance.

This paper proposes a distinct approach to solve sparse-reward multiagent problems.
The inspiration is to reduce the 2nd-sparsity to 1st-sparsity
by introducing prior knowledge about the most fundamental cooperation behaviors.
The answer we put forward is Cooperation Graph Multiagent Reinforcement Learning (CG-MARL).

The proposed algorithm is established revolving around a three-layer Cooperation Graph,
The first layer is the agent layer, each node representing an individual agent.
The middle layer is composed of clusters, the basic unit to execute cooperative actions.
The last layer is the target layer, and each node is a delegate of a cluster-action.
The Cooperation Graph itself does not consist of neural networks,
instead,
in order to achieve dynamic environment responses, it accepts manipulation from a policy neural network, 
which is designed based on attention mechanism \cite{vaswani2017attention} and trained by RL optimizers.
The adjustment actions $\vec{\mu}(t)$ provided by this policy neural network is used to re-map edges of Cooperation Graph.
Clusters parse instruction from targets' cluster-actions $f$ and 
translate them to original-actions $u$ of individual agents.

The performance of CG-MARL is tested with two sparse-reward benchmark environments,
Anti-Invasion Interception (AII) and Hazardous Cargo Transport(HCT).
The result demonstrates the state-of-the-art performance of our model in both environments.

The source code used in this paper is open at the following url: https://github.com/binary-husky/hmp2g.

\section{Methods}
\begin{figure}[!t]
	\centering
	\includegraphics[width=0.8\linewidth]{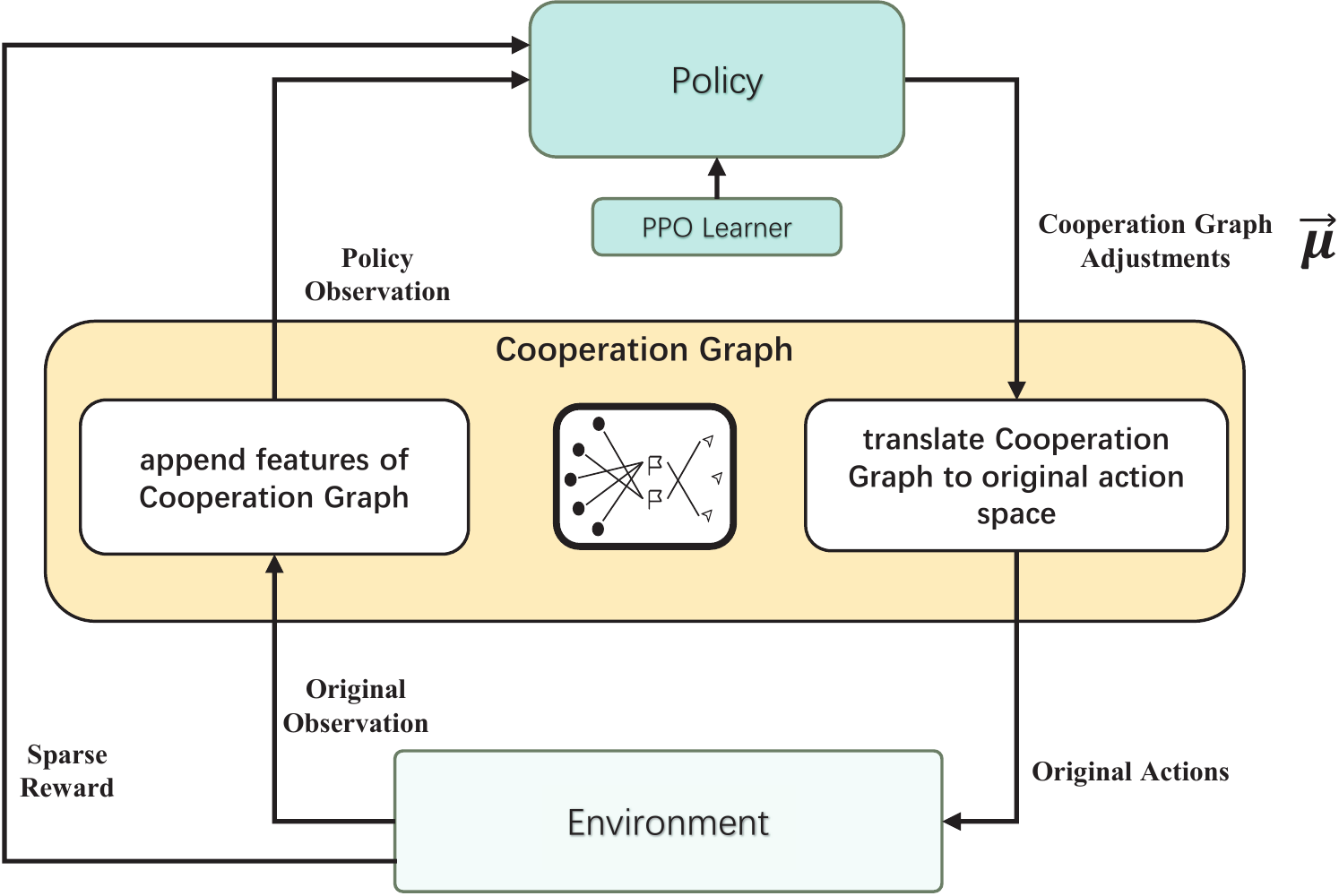}
	\caption{The overall framework of CG-MARL.}
	\label{CG-MARL}
  \end{figure}

In this section,
the framework of the CG-MARL algorithm is introduced.
The overall structure of our algorithm is shown in Fig.~\ref{CG-MARL}.
It has a hierarchical feature considering the relationship between the policy neural network, 
the Cooperation Graph and the original environment.
First, we introduce Cooperation Graph, the core concept of our model that is illustrated as yellow in Fig.~\ref{CG-MARL}.
Second, we explain how a Cooperation Graph interacts with the environment by a procedure called Cooperation Graph Translation.
Then, the interaction between Cooperation Graph and policy neural network is demonstrated in the \textit{Graph Adjustment} section.
Next, the \textit{Attention-based Policy Network} section introduce the policy neural network, which is established from soft attention modules.
Finally, we propose a Time Step Degeneration (TSD) technique to speed up and stabilize the training process.

\subsection{The Basic Structure of Cooperation Graph}
\begin{figure}[!t]
	\centering
	\includegraphics[width=0.8\linewidth]{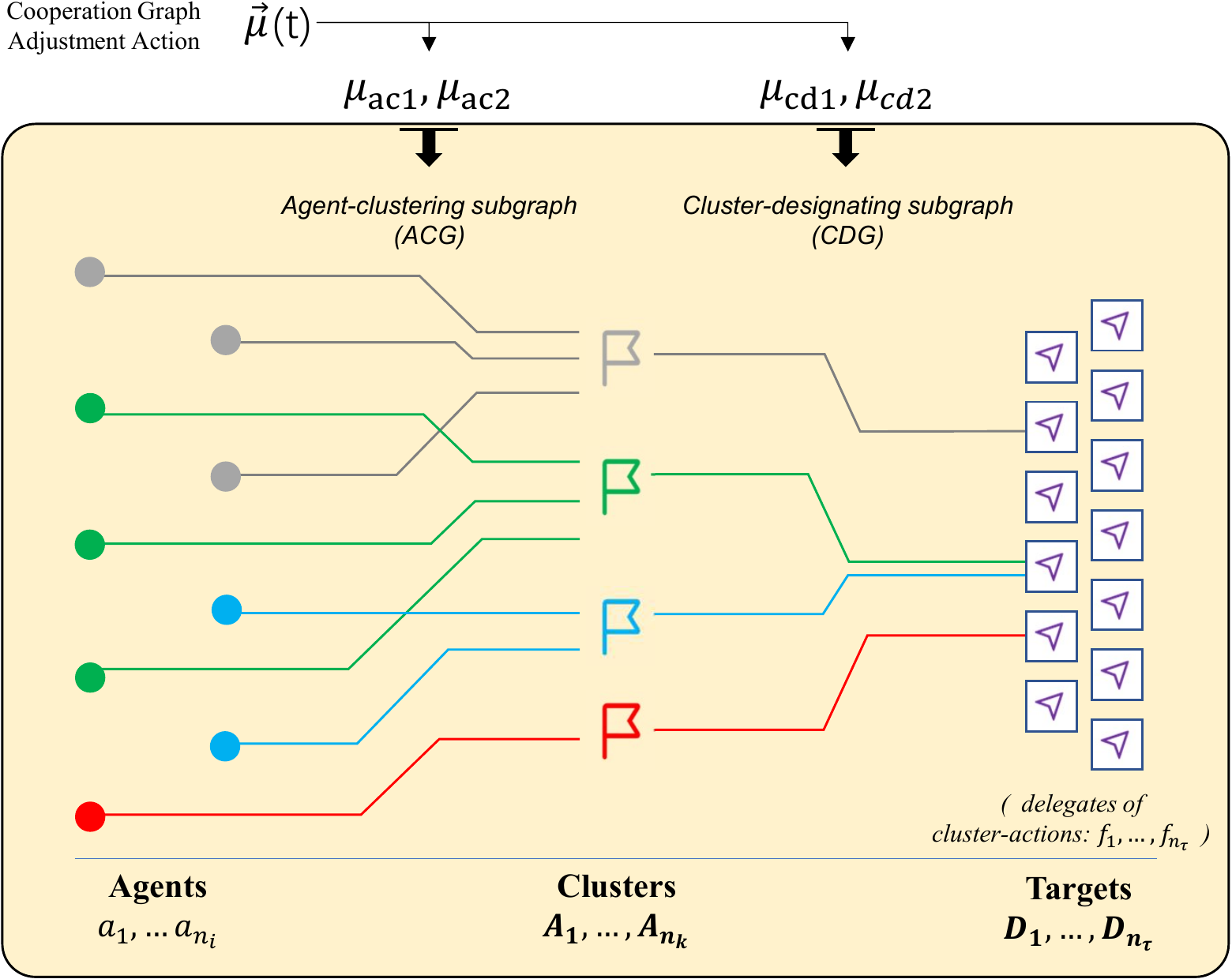}
	\caption{The structure of Cooperation Graph.}
	\label{CG}
	\label{CG-Struct}
  \end{figure}

The Cooperation Graph itself contains no neural network,
instead, it is designed as a simple three-layer directed graph as shown in Fig.~\ref{CG-Struct}
The first layer is the agent layer, 
each node of which represents an individual team agent $a_i \in \mathcal{A}$, 
where $A$ is the collection of all agents.

The second layer is composed of $n_k$ clusters.
Clusters are containers of agents. 
At any moment $t$,
each agent node has only one edge connecting it to a single cluster $A_k(t)$,
indicating that an agent only belongs to a single cluster at a certain time step.
On the other hand, a cluster can hold any number of agents $A_k = \{a_{k(1)}, a_{k(2)},\dots  a_{k(|A_k|)}\}$, 
where $a_{k(*)}\in A_k$ represents an agent linked to this cluster, 
and $|A_k|$ is the number of agents this cluster holds.
This constraint can be formally written as:
$$
A_i(t)\cup A_j(t)=\varnothing, \forall 1 \leq i \neq j \leq n_k 
$$
$$
\mathcal{A} = \bigcup_{k=1}^{n_k}A_k(t)
$$
We refer to a cluster containing no agents as an empty cluster.
Agents are allowed to gather into one single cluster or spread over to create more valid clusters according to the requirements of tasks.

The last layer is a target layer, in which there are $n_\tau$ target nodes.
A target node is denoted $\mathbf{D}_\tau$.
The relationship between \textit{\textbf{targets} and clusters} 
is similar to that of \textit{\textbf{clusters} and agents} in the Cooperation Graph.
A cluster only connects to one target node at each time step,
but a target can contain multiple clusters $\mathbf{D}_\tau=\{A_{\tau(1)},\dots,A_{\tau(|\mathbf{D}_\tau|)}\}$.
This constraint can be formally written as:
$$
\mathbf{D}_i(t)\cup \mathbf{D}_j(t)=\varnothing, \forall 1 \leq i \neq j \leq n_\tau
$$
$$
\mathcal{C} = \bigcup_{\tau=1}^{n_\tau}\mathbf{D}_\tau(t)
$$
where $\mathcal{C}=\{A_1,\dots,A_{n_k}\}$ is the set of all cluster nodes, 
$|\mathbf{D}_\tau|$ is the number of clusters that a target node $\mathbf{D}_\tau$ holds.

The target nodes are delegates of pre-defined cluster-actions $\{f_1, \dots, f_{n_\tau}\}$.
A cluster-action $f_\tau$ is different from a original-action $u_i$ that can be committed directly to the environment.
A original-action only takes effect on a single agent.
By contrast,
a cluster-action controls the behaviors of multiple agents in a cluster simultaneously.

Agent nodes cannot directly connect to target nodes.
Consequently, the cooperation graph can be decomposed as two bipartite subgraphs:
the subgraph between agent and cluster layers is the Agent Clustering subgraph (ACG),
and the subgraph connecting clusters with targets is the Cluster Designating subgraph (CDG).
The clusters are the center of the entire cooperation graph framework.

\subsection{The Translation of Cooperation Graph}
In order to control agents with Cooperation Graph,
we need to translate the nodes and edges of Cooperation Graph 
into actual agent actions in the action space of the environment (original-action).
We refer to this procedure as Cooperation Graph translation.

The translation is performed independently within each cluster node.
If a cluster node $A_k$ is not connected by any agents $A_k=\varnothing$,
no translation is needed.
Otherwise, $|A_k|>0$ and
the cluster immediately get the permission to manipulate its member agents $\{a_{k(1)},\dots, a_{k(|A_k|)}\}$.

Then, the cluster $A_k$ accesses the target node that it connects.
Note that each cluster can only be connected to one target node,
and each target node is a delegate of a cluster-action.
This target node is denoted as $\mathbf{D}_{\tau(k)}$ and the delegated cluster-action as $f_{\tau(k)}$.
$A_k \in \mathbf{D}_{\tau(k)}$.

Last but not the least, the original actions of agents $\{u_{k(1)},\dots, u_{k(|A_k|)}\}$ is obtained by:
$$
U_k(t)=\{u_{k(1)},\dots, u_{k(|A_k|)}\} = f_{\tau(k)}(A_k(t) \mid o(t))
$$
where $o(t)$ is current observation, $U_k$ represents the original-actions of agents in cluster $A_k$.

In this framework, 
there is a high degree of freedom considering the designs of cluster-actions $f_\tau$.
The simplest way is to mirror actions from original-action space directly:
$$
u_{k(1)}=\dots = u_{k(|A_k|)}=f_\tau(t)\in \mathcal{U}
$$
where $\mathcal{U}$ represents the original action space.
Under this setting, agents in a cluster are only allowed to do unified actions,
for instance, making all agents press a button simultaneously.

However, many cooperation patterns cannot be described this way.
As an example, 
if agents' original action space is the velocity (magnitude and direction),
following \textbf{cluster-actions} cannot be achieved through action 
mirroring despite the fact that they are very simple:
\begin{itemize}
	\item a) Gathering at space coordinate $\vec{p}$
	\item b) Gathering at space coordinate $\vec{p}$, forming square formation on the way.
	\item c) Gathering at space coordinate $\vec{p}$, changing speed to reach $\vec{p}$ simultaneously.
\end{itemize}
In this case, instead of mirroring cluster-actions directly from original-action space,
each cluster-action $f_\tau$ function has to consider the observation of each individual agent.
For example, 
assign cluster members with different velocity magnitude (original-action) based on distances.

There are patterns of cooperation that are obvious and simple to humans,
but will cost significant amounts of resources to learn using pure reinforcement learning and reward guidance,
e.g., the case of geese formation that we mentioned in \textit{Introduction}.
Cluster-action in our CG-MARL algorithm can provide us with an extensible interface
for introducing fundamental cooperation knowledge into reinforcement learning.
Cluster-action can be customized differently in various multiagent tasks.
Using this algorithm framework flexibly will significantly reduce the cost 
of learning complex emergent cooperation behaviors.


\subsection{Initialization of the Cooperation Graph}
A static Cooperation Graph cannot solve real MARL problems,
but a dynamic one can.
We use reinforcement learning algorithms are used to construct
policies to manipulate Cooperation Graphs.
Here we introduce our approach to initialize a Cooperation Graph before RL policy gets involved.

In general, 
we randomly initialize the edges of the Cooperation Graph under constraints
described in \textit{Cooperation Graph} section when the training starts.
Then the initial state is stored as fixed parameters.
The initial state of the Cooperation Graph is restored whenever a new episode begins.

In practice, we found that a high-entropy initial state is more likely to increase the speed of training.
The entropy is calculated by:
$$H=-\sum_{k= 1}^{n_k} p(k) \cdot \log p(k), p(k)=\frac{|A_k|}{|\mathcal{A}|}$$
where $H$ is the entropy, $|A_k|$ is the number of agent nodes in $k$-th cluster, $|\mathcal{A}|$ is the total number of agents.
To ensure the Cooperation Graph initialization has higher entropy,
we will randomly generate multiple initial states, 
rank by their entropy and then pick the one with maximum $H$.

\subsection{Cooperation Graph Adjustment}

\begin{figure}[!t]
	\centering
	\includegraphics[width=0.68\linewidth]{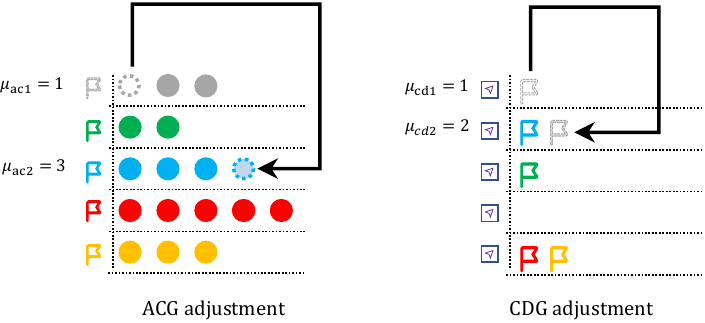}
	\caption{An example of adjustment-actions for Cooperation Graph.}
	\label{adjust}
  \end{figure}
We previously introduced the model of a static Cooperation Graph.
However, 
the Cooperation Graph has to continuously adjust its internal edge connections 
to control the pattern of cooperation under different situations.
This subsection discusses the approach that a Cooperation Graph is adjusted.

The edges in Cooperation Graph is changed by adjustment-actions 
$\vec{\mu}(t)=\left[\mu_{ac1}(t),\mu_{ac2}(t),\mu_{cd1}(t),\mu_{cd2}(t)\right]$,
where $\mu_{ac1},\mu_{ac2} \in \{1,\dots,n_k\}$ and $\mu_{cd1},\mu_{cd2} \in \{1,\dots,n_\tau\}$.
As illustrated in Fig.~\ref{CG}, 
$\mu_{ac1},\mu_{ac2}$ change the edges between agents and clusters (agent-clustering subgraph, ACG),
and $\mu_{cd1},\mu_{cd2}$ adjust the edges between clusters and targets (cluster-designating subgraph, CDG).

\begin{itemize}
	\item ACG: When $\mu_{ac1} \neq \mu_{ac2}$ and $|A_{\mu_{ac1}}|\neq 0$, an agent from $\mu_{ac1}$-th cluster is moved to $\mu_{ac2}$-th cluster.
	Otherwise ACG remains unchanged. The cluster nodes are treated as containers holding agents as elements.
	\item CDG: When $\mu_{cd1} \neq \mu_{cd2}$ and $|D_{\mu_{cd1}}|\neq 0$, a cluster from $\mu_{cd1}$-th target is moved to $\mu_{cd2}$-th target.
	Otherwise CDG remains unchanged. The target nodes are treated as containers holding clusters as elements.
\end{itemize}

Moreover,  when selecting an element (agent in ACG or cluster in CDG) to move among multiple inside a container (cluster in ACG or target in CDG),
a first-in-first-out principle is followed in the procedure above.
In other words,
the first element \textbf{moving in} will also be the first element \textbf{moving out} in a container.

\subsection{Attention-based Policy Network}
\begin{figure}[!t]
	\centering
	\includegraphics[width=\linewidth]{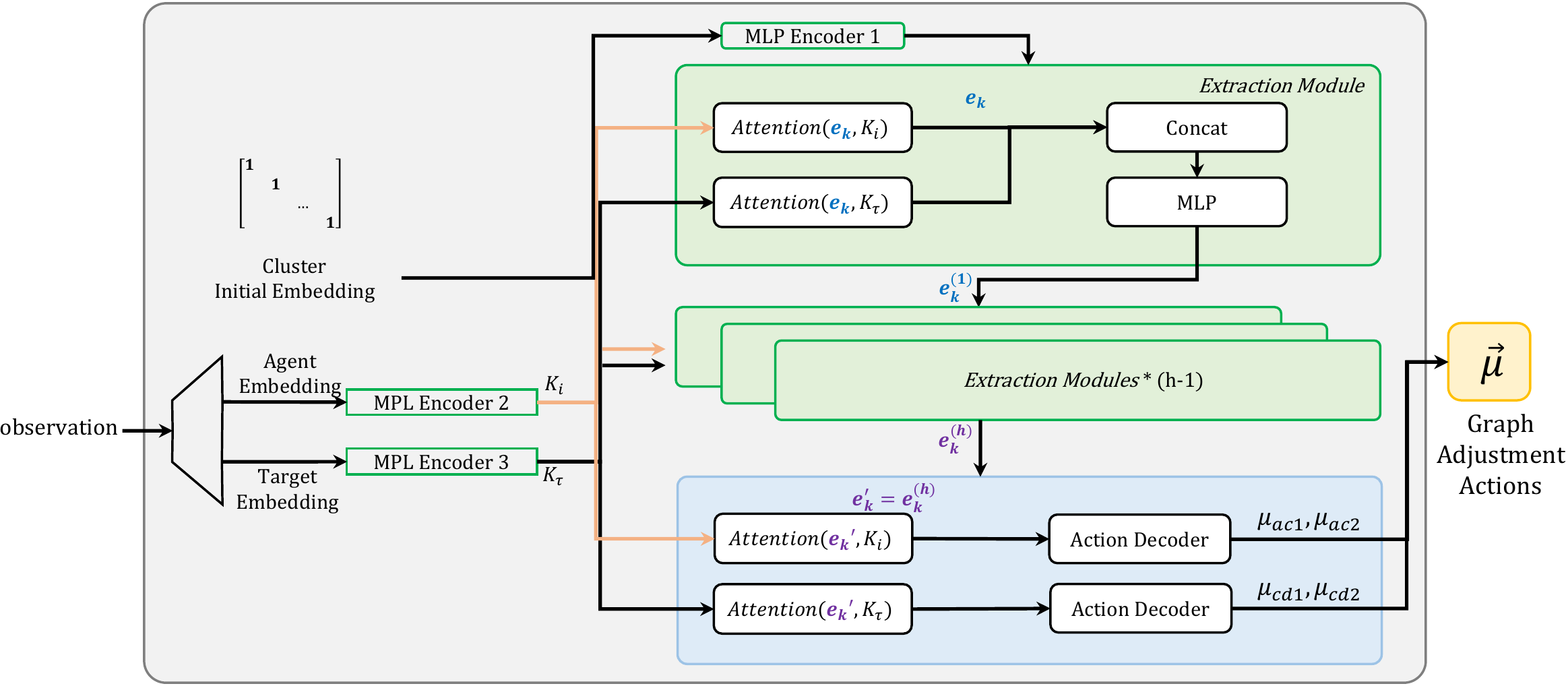}
	\caption{Attention-based Policy Network.}
	\label{network}
  \end{figure}
The attention mechanism is effective in extracting the
relationships between input representations.
Our policy neural network takes advantage of the attention mechanism to learn
the internal relevance between agents, clusters and targets.

As shown in Fig.~\ref{network}, first, the reward signal comes directly from the environment.
Second, in our model, the original observation is reshaped as agent representation $K_i$ and target representation $K_\tau$.
Each cluster is represented by a one-hot vector $e_k \in \mathbb{R}^{n_k}$.
If target observation is not separable, target representation $K_\tau$ can also use one-hot encodings.
Third, $K_i, K_\tau, e_k$ are encoded by three MLP encoder respectively.
Fourth, we update cluster representation by extract feature from $K_i, K_\tau$ with extraction modules,
in which the attention is calculated between $e_k$ and $K_i$, as well as $e_k$ and $K_\tau$.
The outputs is merged by concatenation and MLP-downsampling, producing $e_k^{(1)}$
By stacking $h$ extraction modules and repeating above procedure, $e_k^{(h)}$ can be obtained.
In those extraction module, features about agents and target are highly concentrated into new cluster representations.
Finally, calculate $\operatorname{Attention}(e_k^{(h)}, K_i)$ and $\operatorname{Attention}(e_k^{(h)}, K_\tau)$ 
one more time,
decode the results with fully connected layers to produce the distribution of 
4 graph adjustment actions $\vec{\mu}=\left[\mu_{ac1},\mu_{ac2},\mu_{cd1},\mu_{cd2}\right]$.

PPO \cite{schulman2017proximal} is used to optimize this policy network,
More specifically, we adopted the dual-clip version of PPO proposed in \cite{ye2020mastering}
for robustness.
Our model is efficient enough and needs no tricks on the reward even though the reward is sparse.

\subsection{Time Step Degeneration}
We use a novel Time Step Degeneration (TSD) technique to increase training efficiency.
As a feature of CG-MARL, 
valid graph adjustments are most frequent at the beginning of each episode,
because during this period, the policy neural network needs to prepare Cooperation Graph from its initial state.
In contrast, after Cooperation Graph is prepared, it only needs to be responsive to changes in the environment,
invalid graph adjustments will become frequent, e.g., attempting to move agents between already empty clusters.
If we make adjustment decisions at every single time step, 
valid samples can be flooded by invalid ones.
Consequently, the training efficiency is reduced significantly.

Instead, we use another decision-making sequence to skip time steps that might cause inefficiency.
For example, the graph adjustment is made once every $T_{deg}$ step, 
but the first 5 steps are excluded at the beginning of each episode. 
Reward acquired in skipped steps will be added to the most recent key steps,
thus no reward is missed even though some decision steps are abandoned.

\subsection{Advantage of the Cooperation Graph}

First, the Cooperation Graph successfully reduces the action space requiring consideration and exploration for the policy neural networks. If the original action space is discrete and each agent chooses from action space $\mathcal{U}$, the joint action space is $|\mathcal{U}|^{n_i}$,  which grows exponentially with the number of agents $n_i$. With the implementation of the Cooperation Graph, the new joint action space will be reduced to:
\begin{equation}
	\begin{aligned}
		\vec{\mu} 
		&= \left[\mu_{ac1},\mu_{ac2},\mu_{cd1},\mu_{cd2}\right] \\
		&\in \mathcal{C} \times \mathcal{C} \times \mathcal{T} \times\mathcal{T} =\mathcal{C}^2 \times \mathcal{T}^2	
	\end{aligned}
\end{equation}
where $\mathcal{C}$ is the set of clusters, $n_k = |\mathcal{C}|$ is the number of clusters, 
$\mathcal{T}$ is the set of targets and $n_\tau = |\mathcal{T}|$ is number of the targets (cluster-actions).
This shrink in the policy action space can significantly 
resolve the sparse-reward problem by facilitating exploration efficiency.

Second, the hierarchical design of the Cooperation Graph provides excellent flexibility for introducing prior knowledge about the cooperation task. In practice, this ability helps agents develop behaviors that are difficult or even impossible to learn by policy optimization, for instance, the geese formation problem mentioned in \textit{Introduction}.
Moreover, with an explicit definition of clusters and targets, fundamental cooperative behaviors such as synchronized gathering can be easily achieved by adequately designing cluster-actions.
Our model also has the advantage of working against environmental changes. E.g., after training a model to solve the Anti-Invasion Interception task, if a new constraint is given to forbid agents from moving too far from the protection of landmarks, it is not necessary to train again; instead, add this constraint into 
the code implementation of the related cluster-actions 
and the new requirement will be satisfied immediately.

Third, the agent behaviors learned in CG-MARL are explainable. When needed, a dashboard can be plotted to monitor the edge shifting pattern in the Cooperation Graph. 
And in such a way, we can see which agents are working together and what their current target is.
This characteristic makes it possible to transform a trained CG-MARL model into an interactive model with a human-AI interface. 
In the automated mode, the trained policy neural network is responsible for controlling agents; nevertheless, a human operator can take control of the system anytime by manually rerouting the edges of the Cooperation Graph.
Such human-AI interaction is essential for safety considerations in production environments.
\section{Experiments}
The proposed method is tested on two sparse-reward multi-agent (SRMA) environments, 
Anti-invasion Interception and Hazardous Cargo Transport.
The simulation is carried out in continuous 3-dimensional space. 
All movable objects obey second-order kinematics.
The original-action space is the direction of acceleration $u_i=(acc_x,acc_y,acc_z)$,
where $\|u_i\|=1$.
If an agent commits $\|u_i\| > 1$, $u_i$ will be normalized compulsorily.

\subsection{Anti-Invasion Interception}
\begin{figure}[!t]
	\centering
	\includegraphics[width=\linewidth]{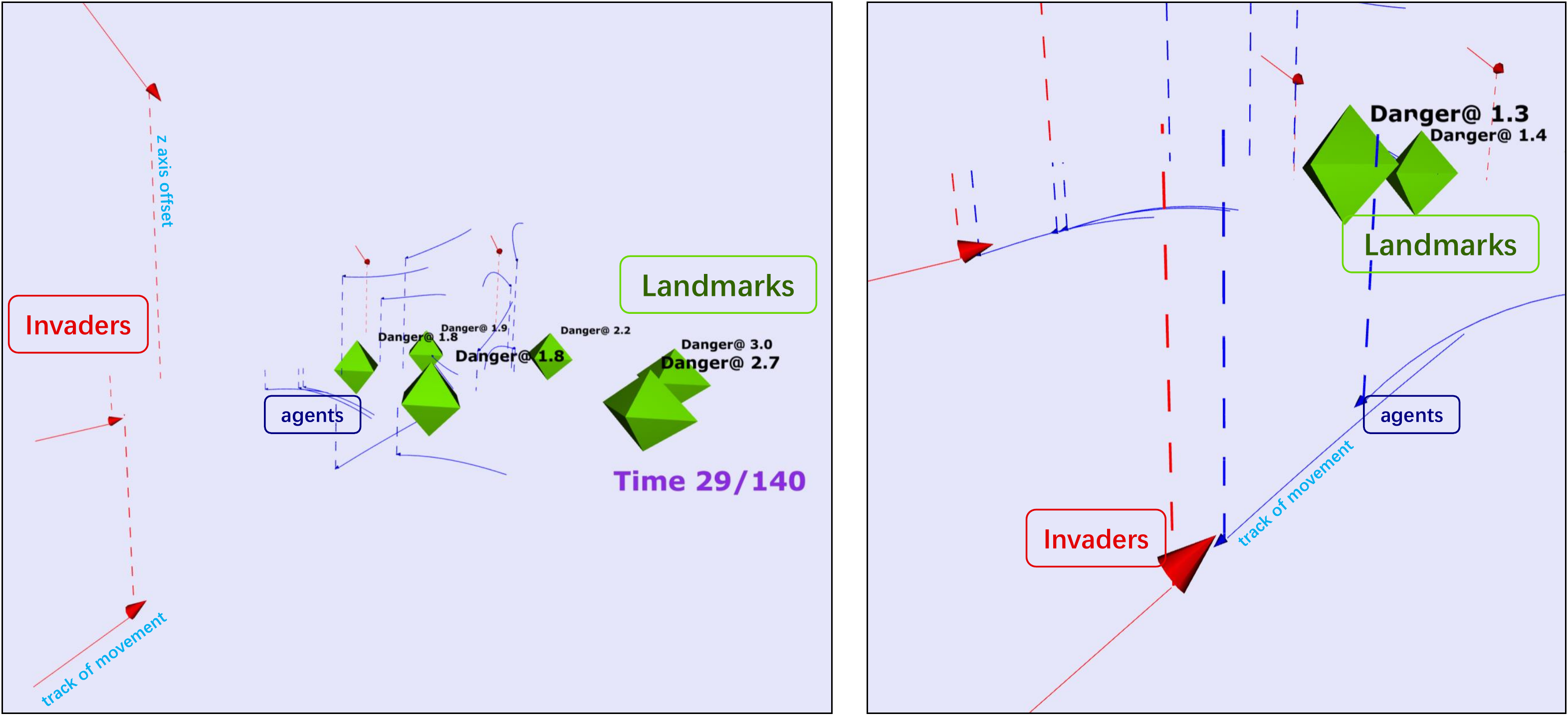}
	\caption{Anti-Invasion Interception (AII) test environment.}
	\label{AII}
  \end{figure}

As illustrated in Fig.~\ref{AII}, 
Anti-Invasion Interception (AII) environment simulates a scenario of $n_A$ agents defending $n_L$ landmarks from $n_I$ invaders.
Invaders are spawned randomly in 3D space when an episode begins,
then head directly towards the nearest landmark for a synchronized invasion. 
If any invaders reach any landmarks, the episode fails.

Agents and invaders share the same maximum velocity.
Agents need to reach and attach themselves to invaders, 
pushing them in the opposite direction with a small force to protect landmarks.
However, agents are much weaker than invaders. 
It takes the joint force of 2 agents to balance the force of an invader.
And it takes at least $3$ agents to reverse the acceleration of an invader, reversing its velocity direction.
An individual agent does not even have the strength of slowing an invader down.
Furthermore, $n_L \geqslant n_I=n_A/3$, 
where $n_L$ is the number of landmarks,
$n_I$ is the number of invaders and $n_i$ is the number of agents.
Under this constraint, agents cannot passively stick to landmarks to succeed.
In this experiment, we select $n_L=6$, $n_I=5$ and $n_A=15$.

The reward setting is extremely straightforward and sparse.
When all landmarks survive through the time limit due to agents' close cooperation,
the environment ends with reward +1, otherwise with reward -1.

\subsection{Hazardous Cargo Transport}

\begin{figure}[!ht]
	\centering
	\includegraphics[width=0.8\linewidth]{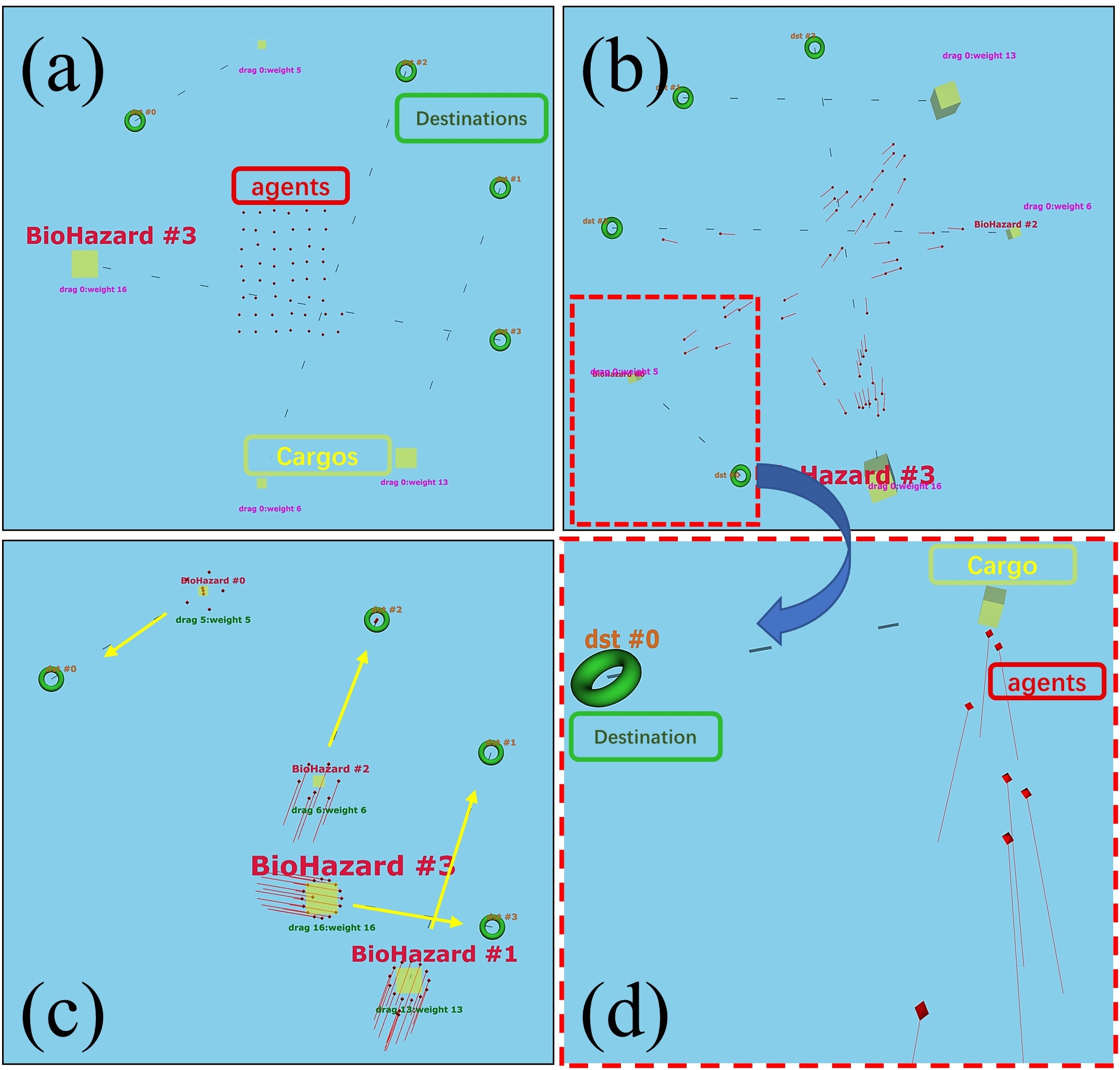}
	\caption{Hazardous Cargo Transport (HCT) test environment. (a) The initial state of HCT. 
	(b) After consideration of the weight of each hazardous cargo, 
	agents head to the position of cargos.
	(c) Agents escorting cargos to safe areas.
	(d) Zooming in (b).}
	\label{cargo}
  \end{figure}

In hazardous cargo transport problems, 
a team with $n_A$ ant-like robot agents need to deliver $n_C$ hazardous cargos 
that weigh much more than themselves to different safe areas within a limited time, 
as it is shown in Figure \ref{cargo}.
If any hazardous cargo fail to reach its corrisponding destination, 
it leaks, resulting in the failure of the task.

When each episode begins, 
each cargo $j$ is assigned with a random weight $w_j$ and a destination $\vec{d}_j$.
The cargo starts to move along with the attached agents when the number of attached agents surpasses its weight. 
And the direction of a cargo's movement is determined by the joint force of agents attached to it,
thus, a cargo may move in the wrong direction if some of the attached agents accelerate towards the wrong way.

The number of total agents $n_A$ is only slightly bigger than the sum of cargo weight, 
$ \Sigma_{j=1}^{p} w_j  = \eta \cdot n_A,  \eta \ge 0.8$. 
Furthermore, robot agents are destroyed along with escorted cargos when reaching their destination to avoid contamination.
From the perspective of agents, achieving a clear division of labor is necessary.
Agents must cooperate and determine careful assignments considering the weight of each cargo.
Otherwise, the team is unable to make ends meet since the number of agents is too limited.
In this experiment we use $n_A=50$ and $n_C=4$.

Agents (as a team) only receive a +0.1 reward when any cargo starts moving, 
and another +0.1 reward when any cargo reaches its destination. 
If all cargos are delivered in time, agents receive another +1 reward.
Overall, agents are required to learn a highly cooperative policy guided by very sparse rewards.
The most difficult problem is the allocation of agents for each cargo,
a mistake of just a few agents can lead to the failure of the entire task.
Also, agents have to solve the destination interoperation problem for each cargo to complete the whole escort process.

\subsection{Cluster-actions and Hyper-parameters}
In both experiment we select the number of cluster nodes $n_k=6$, learning rate $lr=10^{-4}$.
The policy network is updated using an on-policy style,
using samples of the most recent 512 episodes to calculate the policy gradient.

In the AII environment, we design 3 types of cluster-actions:
\begin{itemize}
    \item Gathering at space coordinate selected from a pre-defined list, e.g., $\vec{0}$.
    \item Heading to and attaching to a landmark.
    \item Heading to and attaching to an invader.
\end{itemize}

In the HTC environment, we define 3 types of cluster-actions:
\begin{itemize}
    \item Gathering at space coordinate selected from a pre-defined list.
    \item Heading to and attaching to a hazardous cargo.
    \item Heading to a safe area.
\end{itemize}
\begin{table}[!t]  
	\centering
	\begin{tabular}{ll}  
	\toprule   
	hyperparameter & value   \\  
	\midrule    
	discount factor $\gamma$ & 0.99   \\
	GAE  $\lambda$ & 0.95      \\
	Entropy coefficient   & 0.05      \\
	Gradient clipping  & 0.5      \\
	Number of episodes for each batch  & 512    \\
	PPO epochs $N_{ppo}$ & 16    \\
	Learning rate & $5\cdot 10^{-4}$      \\  
	\bottomrule  
	\end{tabular}
	\caption{Hyper-parameters in PPO optimization, 
	most parameters are the default values widely used in RL literature, 
	and the episode batch size is larger than usual to provide training stability.}
	\label{tb:o-parameters}
\end{table}

The training is performed on a server with RTX3090 GPUs, 
each experiment only uses less than 8G GPU memory.
The training process completes within a day.
Hyper-parameters of PPO algorithm is listed in Table.~\ref{tb:o-parameters}.

\section{Results}

\subsection{Results on AII benchmark}
\begin{figure}[t]
	\centering
	\begin{subfigure}[t]{0.7\linewidth}
	\centering
	\includegraphics[width=.999\linewidth]{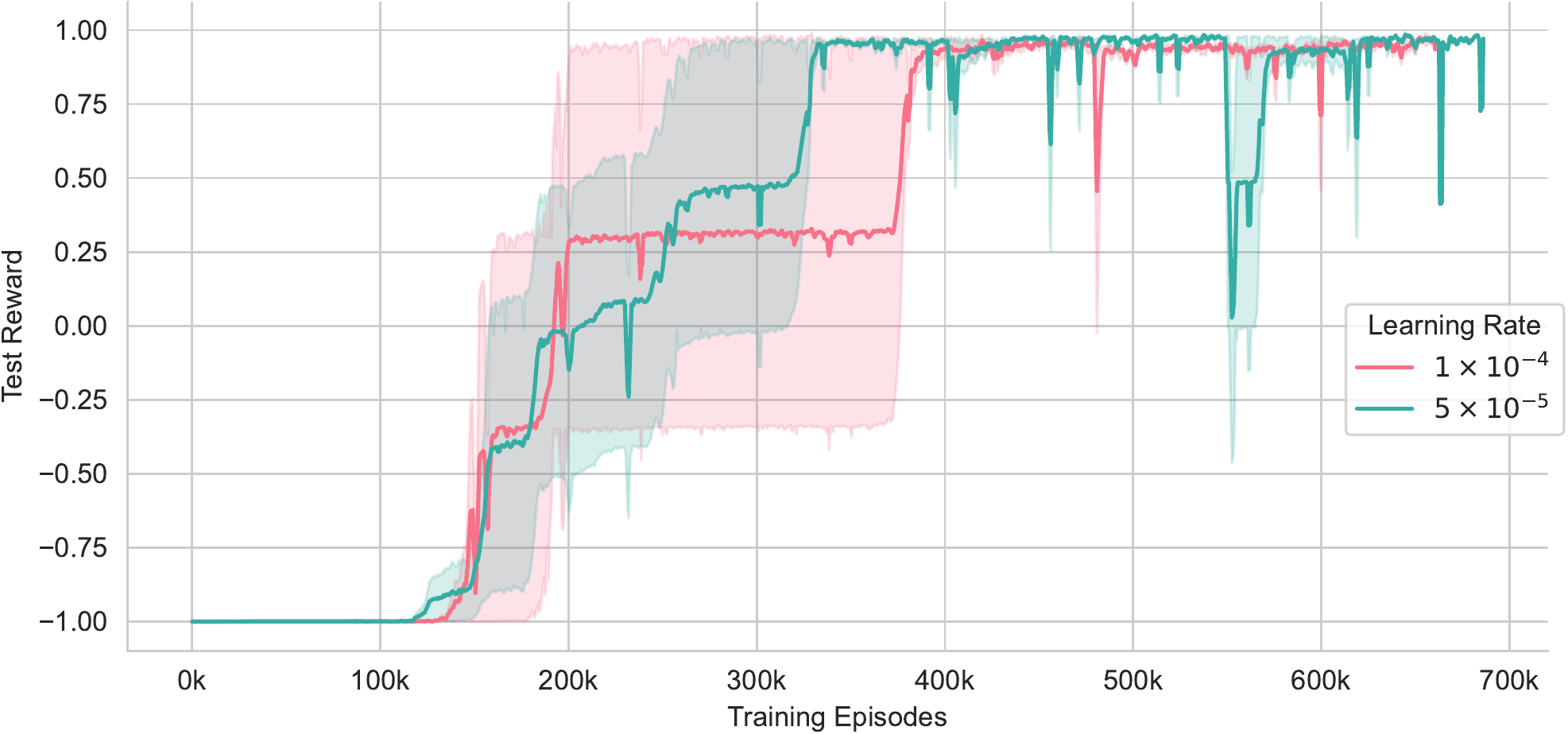}
	\caption{Test reward.}
	\label{fig:aii1}
	\end{subfigure} 
	\\
  \begin{subfigure}[t]{0.7\linewidth}
	\centering
	\includegraphics[width=.999\linewidth]{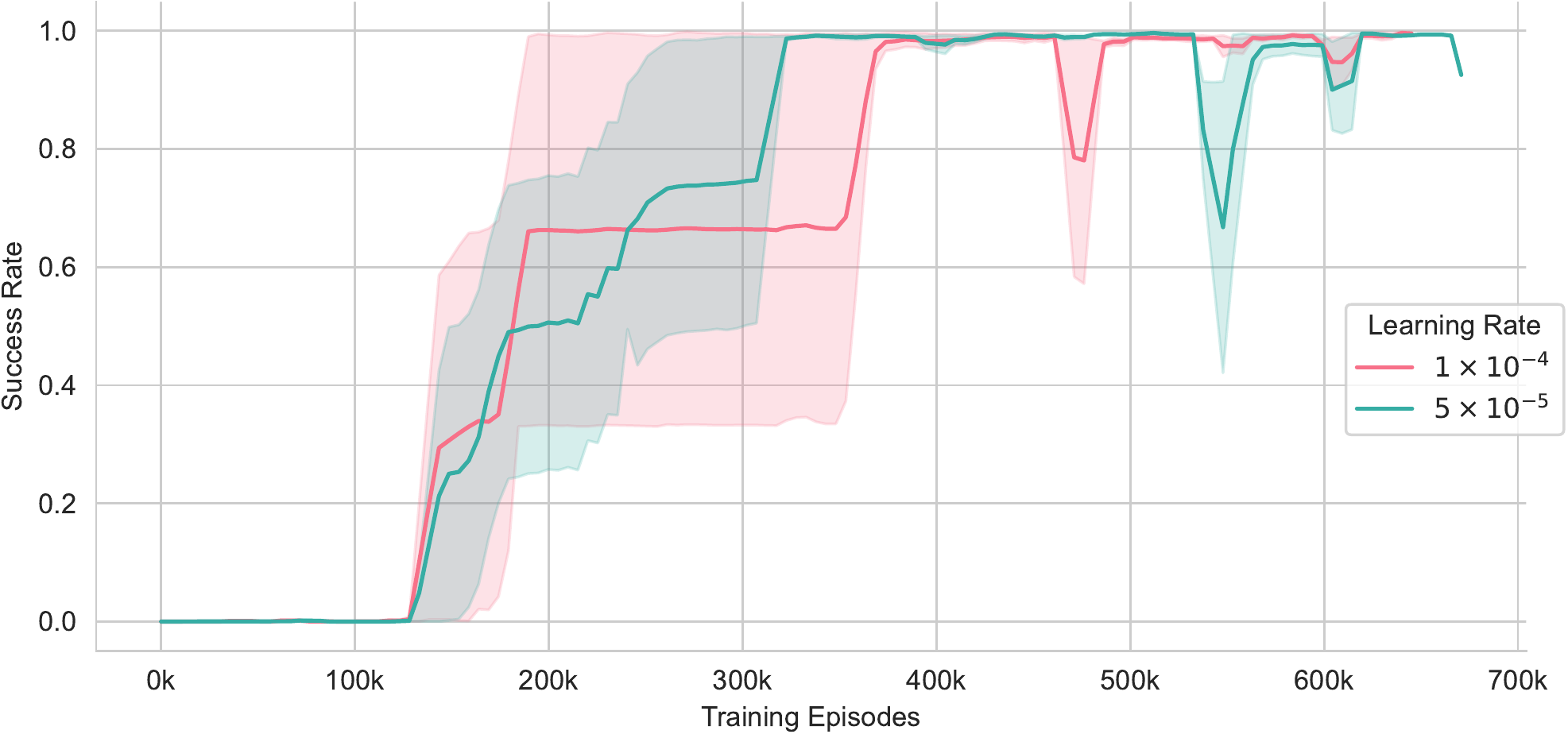}
	\caption{Success rate.}
	\label{fig:aii2}
	\end{subfigure} 
  
  \caption{The performance of CG-MARL in AII benchmark. In this experiment $n_k=6$ and $T_{deg}=5$.}
  \label{fig:aii_rate}
  \end{figure}

Unlike other multiagent benchmarks in existing works, 
AII test environment needs close collaboration of every single agent.
A failure of even one agent leads to the failure of the whole task.
Furthermore, the exploration is very tough 
because only the positive reward is given only when the whole task succeeds.
As it is shown in Fig.~\ref{fig:aii_rate},
agents at the early stage have little chance to get any reward except the -1 failure penalty ($\leq 100$k episode).
Nevertheless,
our model is able to step out of the sparse-reward trap within 200k episodes.

When the learning rate is $1\times 10^{-4}$,
fig.~\ref{fig:aii_rate} also illustrates that the number of episodes needed to reach the top reward ranges from 200k to 400k.
It suggests that the training process is not robust enough.
Therefore, 
we attempted to reduce the learning rate of PPO and find it effective
in stabilizing the policy optimization process.

\subsection{Results on HCT benchmark}
The HCT benchmark is a difficult sparse-reward task because it requires a more precise allocation of agents,
and the fault tolerance is very low.
Therefore, in the HCT environment, we can better investigate the influence of two core hyper-parameters
of CG-MARL, namely the number of cluster nodes $n_k$ in CG and 
TSD degeneration parameter $T_{deg}$.

\begin{figure}[ht]
	\centering
	\begin{subfigure}[t]{0.7\linewidth}
	\centering
	\includegraphics[width=.999\linewidth]{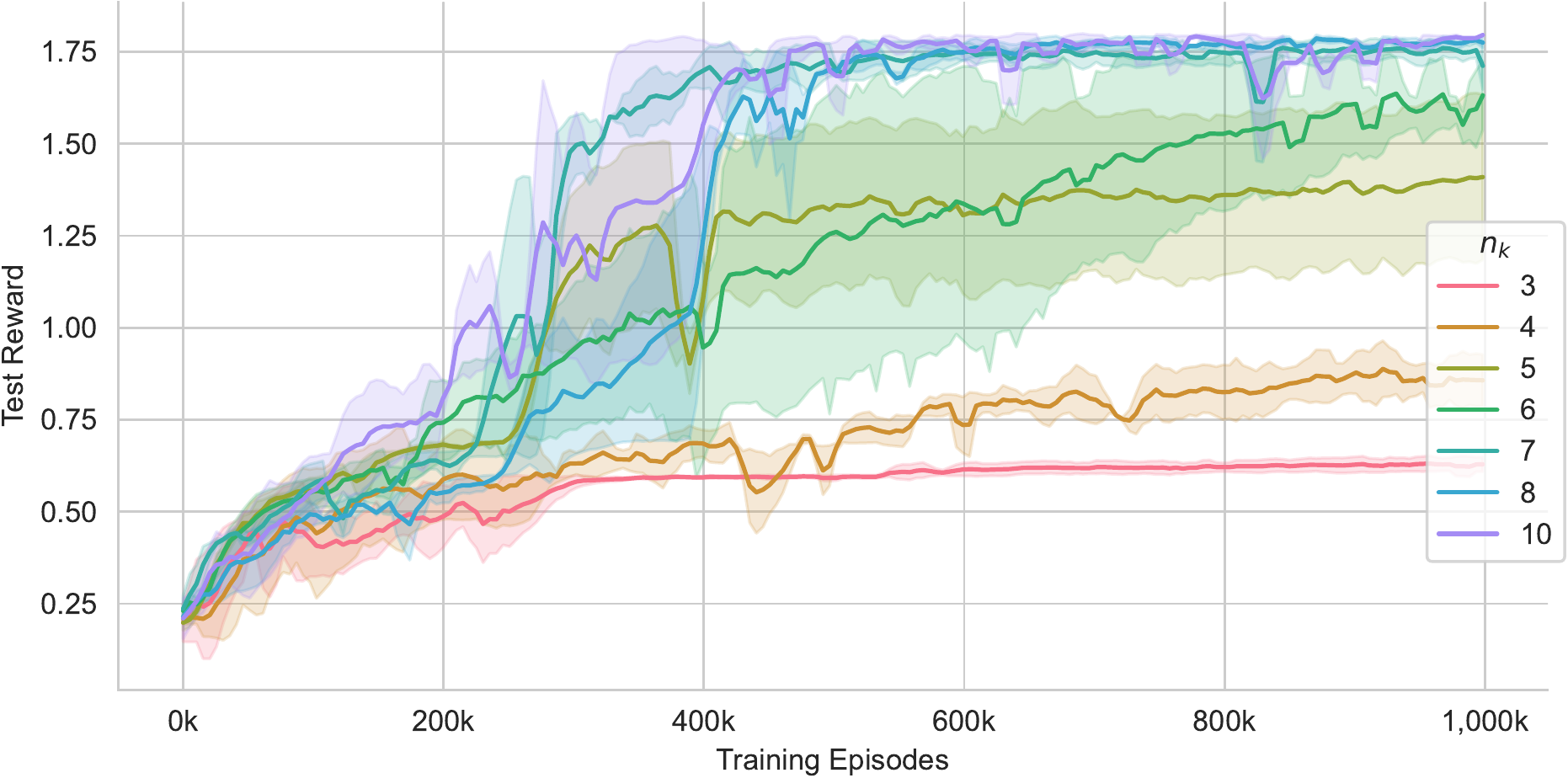}
	\caption{Test reward.}
	\label{fig:dcadeepin1}
	\end{subfigure} 
	\\
  \begin{subfigure}[t]{0.7\linewidth}
	\centering
	\includegraphics[width=.999\linewidth]{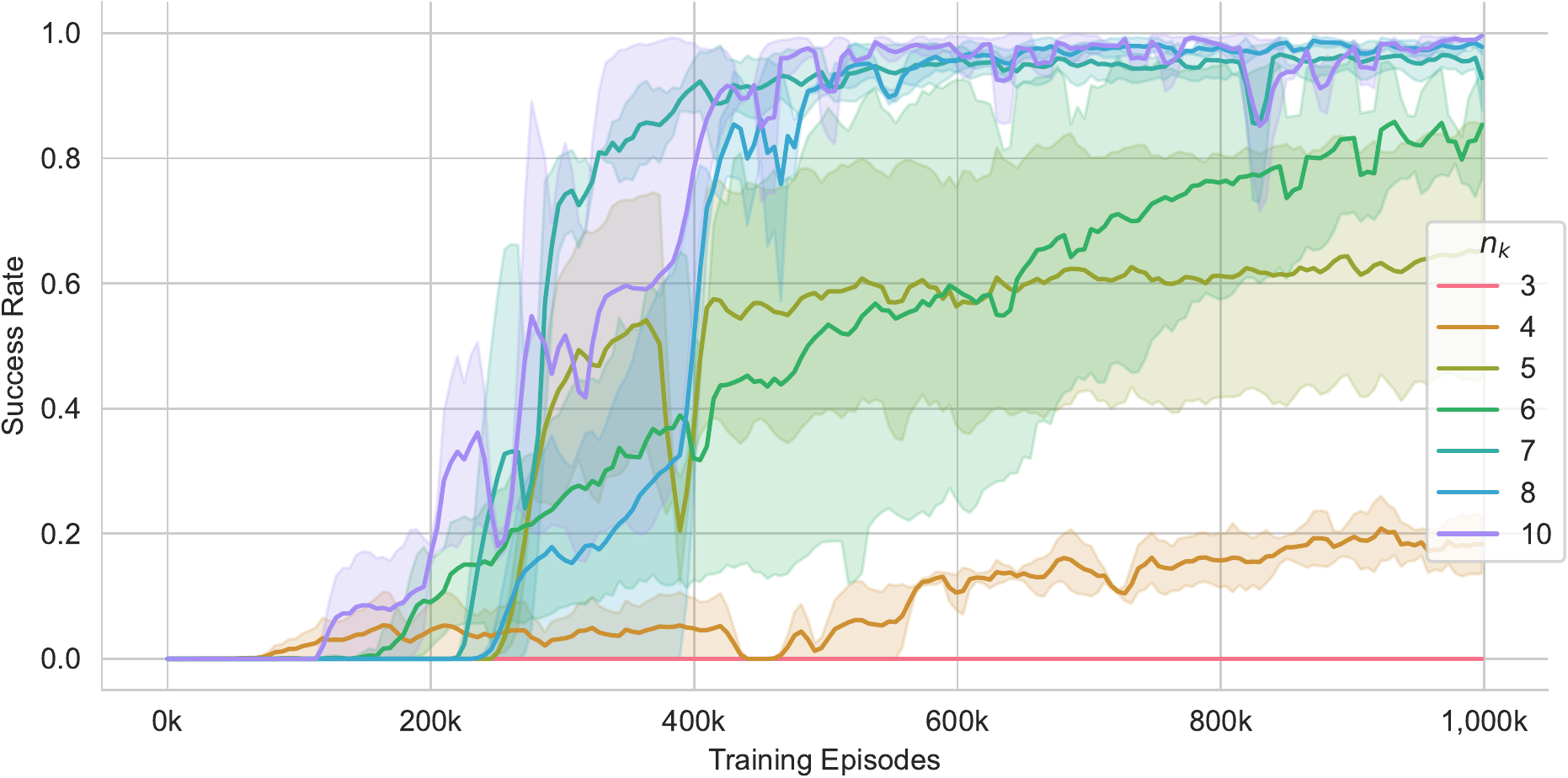}
	\caption{Success rate.}
	\label{fig:dcadeepin3}
	\end{subfigure} 
  
  \caption{The influence of $n_k$ selection in HCT test environment. In these ablation experiments $T_{deg}=25$.}
  \label{fig:hct_rewards}
  \end{figure}

Fig.~\ref{fig:hct_rewards} and Table.~\ref{tab:hct_rewards} illustrates 
that a too small $n_k$ is detrimental to model performance.
When $n_k \leq 4$, 
there are not enough cluster nodes to operate synchronously in order to complete the escort in time.
It can be observed from Table.~\ref{tab:hct_rewards} that 
the success rate rises to 96.58\% when $n_k=6$,
and can be further improved by increasing more cluster nodes.
However, it does not indicate that a large $n_k$ is better 
because a large $n_k$ will significantly increase the scale of the Cooperation Graph and slow down the speed of training.

\begin{figure}[ht]
\centering
\begin{subfigure}[t]{0.7\linewidth}
\centering
\includegraphics[width=.999\linewidth]{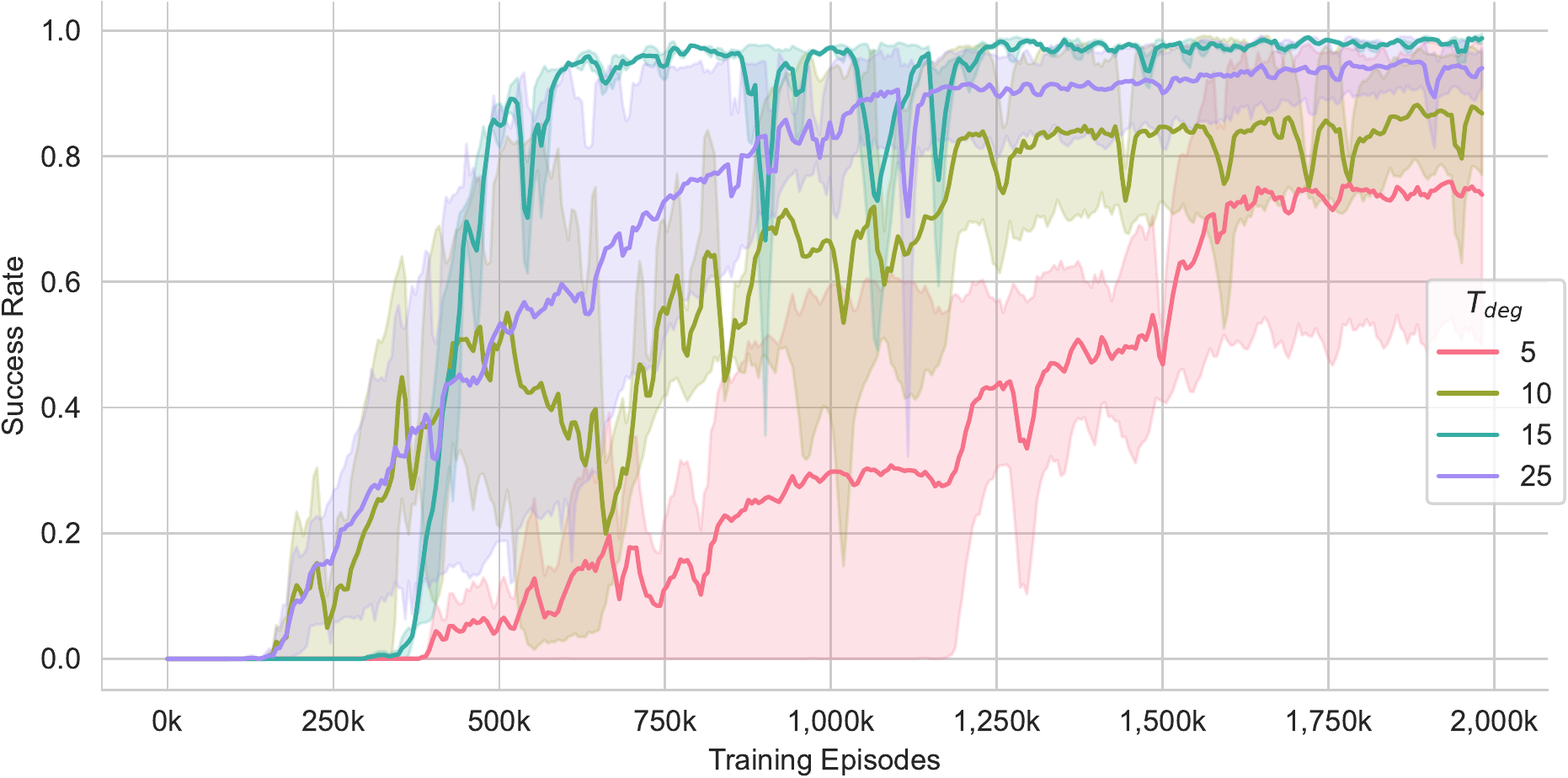}
\caption{Test reward.}
\label{fig:dcadeepin1}
\end{subfigure} 
\caption{The influence of $T_{deg}$ selection in HCT test environment. In these experiments $n_k=6$.}
\label{fig:hct_tdeg_rate}
\end{figure}

During training, we discovered that
applying the Time Step Degeneration (TSD) technique reduces GPU memory consumption.
TSD greatly stabilizes and accelerates the training process by increasing episode batch size to reduce sample variance.
In Fig.~\ref{fig:hct_tdeg_rate}, 
it is noticed that a larger $T_{deg}$ guarantees a higher success rate at the initial stage of training ($\leq$250k episodes).
However, Table.~\ref{tab:hct_rewards} also reveals the side effects of inadequate TSD settings.
Recall that the adjustment of CG takes place every $T_{deg}$ environment steps.
$T_{deg}=1$ means CG is adjusted on every step and TSD is disabled.
Increasing $T_{deg}$ indicates that policy neural network becomes \text{lazier}.
When $T_{deg}<15$, the model performance benefits from this laziness.
In contrast, when $T_{deg}>15$, the CG becomes more inactive and less responsive.
Consequently, the performance reduced correspondingly.

\begin{table}
	\centering
	\setlength{\tabcolsep}{1mm}{
	\begin{tabular}{c|c|ccccc} 
	\hline
	\multicolumn{2}{c|}{$n_k$, fix $T_{deg}$=25}                       & 3      & 4      & 6                        & 8      & 10               \\ 
	\hline
	\multirow{2}{*}{Test Reward}  & mean & 0.6398 & 0.9588 & 1.7609                   & 1.7941 & \textbf{1.7979}  \\ 
	\cline{2-7}
																					 & std  & 0.0373 & 0.0970 & 0.0481                   & 0.0083 & 0.0030           \\ 
	\hline
	\multirow{2}{*}{Success Rate} & mean & 0.0000 & 0.2676 & 0.9658                   & 0.9951 & \textbf{0.9980}  \\ 
	\cline{2-7}
																					 & std  & 0.0000 & 0.0691 & 0.0426                   & 0.0069 & 0.0028           \\ 
	\hline
	\multicolumn{7}{c}{}                                                                                                                                             \\ 
	\hline
	\multicolumn{2}{c|}{$T_{deg}$, fix $n_k$=6}                         & 5      & 10     & 15                       & 25     &          -        \\ 
	\hline
	\multirow{2}{*}{Test Reward}  & mean & 1.5543 & 1.6952 & \textbf{1.7966} & 1.7609 &         -         \\ 
	\cline{2-7}
																					 & std  & 0.3287 & 0.1449 & 0.0048                   & 0.0481 &         -         \\ 
	\hline
	\multirow{2}{*}{Success Rate} & mean & 0.7793 & 0.9062 & \textbf{0.9971} & 0.9658 &        -          \\ 
	\cline{2-7}
																					 & std  & 0.2955 & 0.1298 & 0.0041                   & 0.0426 &         -         \\
	\hline
	\end{tabular}}
	\caption{The ablation study of the cluster node number $n_k$ and TSD degeneration parameter $T_{deg}$ in HCT environment.}
  	\label{tab:hct_rewards}
\end{table}

\subsection{Limitations of CG-MARL}
Despite the success of the proposed CG-MARL model in many sparse-reward problems, we also noticed many limitations of our model.
First, the introduction of the Cooperation Graph comes with too many extra hyper-parameters, such as the number of clusters $n_k$ and the TSD parameter $T_{deg}$.
Although we have provided the principle of selecting these parameters via ablation experiments, they still cause an inconvenience when assigned to solve a new task.
Second, when the number of agents surges to hundreds, 
the Cooperation Graph will need to transfer more agents between clusters during a fixed period of time. 
In such cases, the TSD method used in our model will be influenced, resulting in performance decay.


\section{Conclusions}
We have proposed a Cooperation Graph Multiagent Reinforcement Learning algorithm (CG-MARL).
This novel algorithm provides excellent efficiency for solving sparse-reward multiagent tasks.
CG-MARL introduces a novel structure, namely the Cooperation Graph (CG),
into the framework of reinforcement learning.
CG-MARL algorithm successfully combines the emergent agent behaviors with fundamental cooperation knowledge,
providing a low-cost approach to finish highly cooperative tasks.
The experiments demonstrate that our model has state-of-the-art performance 
in sparse-reward multiagent benchmarks including AII and HCT,
which directly hit the weakness of existing MARL methods.
In ablation experiments, 
we demonstrate and explain the function of two core parameters of CG-MARL.
Our algorithm still has many limitations. 
We have not discovered general-purpose cluster-actions nor the way to design them automatically.
We expect that our work can inspire further studies of sparse-reward MARL.

\section{Acknowledgments}
This work was supported in part by 
the National Key Research and Development Program of China (2018AAA0102404), 
the National Natural Science Foundation of China (62073323), 
the Strategic Priority Research Program of Chinese Academy of Sciences (XDA27030204),
the External Cooperation Key Project of Chinese Academy Sciences (173211KYSB20200002),
and 
the Science and Technology Development Fund of Macau (No.0025/2019/AKP).


\bibliographystyle{IEEEtran.bst}
\bibliography{cite.bib}


\end{document}